\pdfoutput=1

\documentclass[11pt]{article}

\usepackage[]{EMNLP2022}

\usepackage{times}
\usepackage{latexsym}

\usepackage[T1]{fontenc}

\usepackage[utf8]{inputenc}

\usepackage{microtype}

\usepackage{inconsolata}

\usepackage{url}

\usepackage{booktabs}
\usepackage{multirow}
\usepackage{subcaption}

\usepackage{amssymb} 
\usepackage{amsmath}

\usepackage{CJKutf8}

\usepackage{graphbox}
\usepackage{xurl}

\usepackage{tikz}
\usepackage{tikz-qtree}
\usetikzlibrary{arrows,decorations.pathmorphing,backgrounds,positioning,fit,petri,shapes.misc, arrows.meta,shapes.geometric,decorations.markings,calc,shadows.blur,decorations.pathreplacing,quotes}
\definecolor{myblue}{RGB}{6, 82, 221}
\definecolor{myorange}{RGB}{211, 84, 0}
\definecolor{lowblue}{RGB}{102,178,255}
\definecolor{justblue}{RGB}{84, 160, 255}
\definecolor{mypurple}{RGB}{108, 92, 231}
\definecolor{mygray}{RGB}{158, 158, 158}
\definecolor{lowpurple}{RGB}{204,153,255}
\definecolor{lowwhite}{RGB}{255,255,255}
\definecolor{verylowpurple}{RGB}{255,102,102}
\definecolor{embcolor}{RGB}{255,255,255}
\definecolor{myred}{RGB}{235, 47, 6} 
\definecolor{mygreen}{RGB}{162, 217, 206} 
\definecolor{fontgrey}{RGB}{44, 62, 80}
\definecolor{lowpurple}{RGB}{210, 180, 222}
\definecolor{mypumpkin}{RGB}{229, 152, 102}
\definecolor{lowgreen}{RGB}{171, 235, 198}
\definecolor{lowgreen2}{RGB}{186, 220, 88}
\definecolor{lowred}{RGB}{245, 183, 177}
\definecolor{lowyellow}{RGB}{241, 196, 15}
\definecolor{mypink}{RGB}{255, 118, 117}
\definecolor{bluemartina}{RGB}{18, 203, 196}
\definecolor{puffin}{RGB}{250, 152, 58}
\definecolor{grass}{RGB}{0, 148, 50}
\definecolor{cnngray}{RGB}{116, 125, 140}

\newcommand{\squishlist}{
	\begin{list}{$\bullet$}
		{ \setlength{\itemsep}{0pt}
			\setlength{\parsep}{3pt}
			\setlength{\topsep}{3pt}
			\setlength{\partopsep}{0pt}
			\setlength{\leftmargin}{1.5em}
			\setlength{\labelwidth}{1em}
			\setlength{\labelsep}{0.5em} } }

	\newcounter{Lcount}
	\newcommand{\squishlisttwo}{
		\begin{list}{\arabic{Lcount}. }
			{ \usecounter{Lcount}
				\setlength{\itemsep}{0pt}
				\setlength{\parsep}{0pt}
				\setlength{\topsep}{0pt}
				\setlength{\partopsep}{0pt}
				\setlength{\leftmargin}{2em}
				\setlength{\labelwidth}{1.5em}
				\setlength{\labelsep}{0.5em} } }
		
		\newcommand{\squishend}{
	\end{list} }

\usepackage{multirow}
\usepackage{hhline}
\usepackage{tabularx}
\usepackage{ragged2e}
\newcolumntype{Y}{>{\RaggedRight\let\newline\\\arraybackslash\hspace{0pt}}X} 
\usepackage{xcolor}
\usepackage{pgfplots, pgfplotstable}
\pgfplotsset{compat=1.17} 
\usepackage{pgfpages}
\usepackage{mathbbol}
\usepackage{arydshln}
\usepackage{graphicx}

\usepackage{amssymb}
\usepackage{amsfonts}

\usepackage{algorithm}
\usepackage{algorithmic}
\usepackage{pgf}

\usepackage{lipsum}
\usepackage{multicol}
\usepackage{changepage}

\usepackage[export]{adjustbox}

\usepackage[utf8]{inputenc}
 
\usepackage{amsthm}

\usepackage{flushend}

\title{SentBS: Sentence-level Beam Search for Controllable Summarization}

\author{
\textbf{
Chenhui Shen \thanks{~~Chenhui and Liying are under the Joint PhD Program between Alibaba and their corresponding universities.}~~\textsuperscript{\rm 1,2}~~
Liying Cheng  $^*$\textsuperscript{\rm 1,3}~~ 
Lidong Bing\thanks{$^\dag$ Corresponding author.}$^\dag$\textsuperscript{\rm 1}~~ 
Yang You\textsuperscript{\rm 2}~~
Luo Si\textsuperscript{\rm 1}}\\
\textsuperscript{\rm 1}DAMO Academy, Alibaba Group~~
\textsuperscript{\rm 2} National University of Singapore\\
\textsuperscript{\rm 3}Singapore University of Technology and Design ~~
\\
{\tt\{chenhui.shen, liying.cheng\}@alibaba-inc.com} \\
{\tt\{l.bing, luo.si\}@alibaba-inc.com}
~~{\tt youy@comp.nus.edu.sg}}

\begin{document}
\maketitle
\begin{abstract}
A wide range of control perspectives have been explored in controllable text generation. 
Structure-controlled summarization is recently proposed 
as a useful and interesting research direction.
However, current structure-controlling methods 
have limited effectiveness in enforcing
the desired structure.
To address this limitation, we 
propose a sentence-level beam search generation method (SentBS), 
where evaluation is conducted throughout the generation process to select suitable sentences for subsequent generations.
We experiment with different combinations of decoding methods to be used as sub-components by SentBS and evaluate results on the structure-controlled dataset MReD.
Experiments show that all explored combinations for SentBS can improve the agreement between the generated text and the desired structure, 
with the best method significantly reducing the structural discrepancies suffered by the existing model, by approximately 68\%.
\footnote{Our code and data are available at \url{https://github.com/Shen-Chenhui/SentBS}.} 
\end{abstract}

\begin{figure*}[t!]
    \begin{center}
    \resizebox{0.85\linewidth}{!}{
    \includegraphics[width=\textwidth]
    {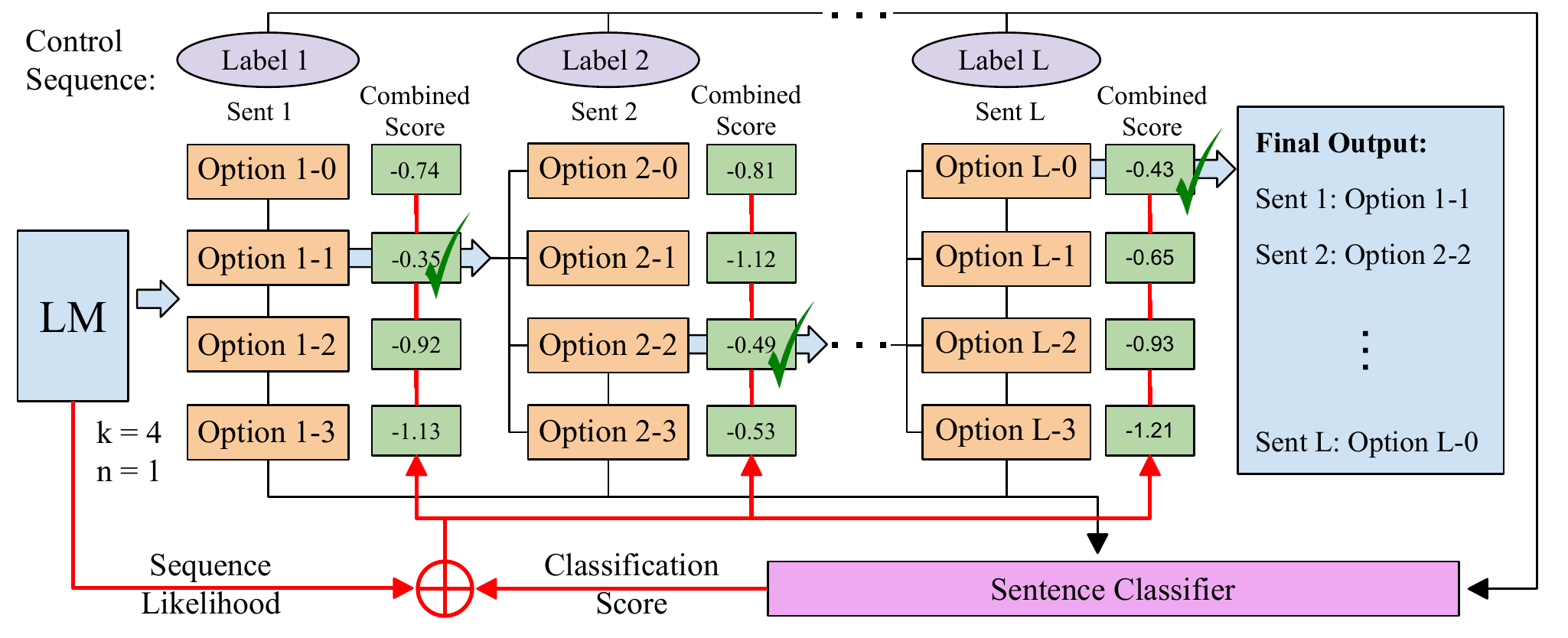}}
    \end{center}
    \vspace{-4mm}
    \caption{Illustration of SentBS.
    The score values are for illustration purposes only. For simplicity, we only illustrate for $k=4$ and $n=1$.}
    \label{fig:sentbs}
    \vspace{-6mm}
\end{figure*}

\section{Introduction}






Controllable text generation is receiving increasing attention due to its wide range of applications. 
Depending on the use cases, the controllable generation tasks may focus on a wide range of control perspectives, such as
entities \cite{narayan2022well, fan2018controllable}, aspects \cite{hayashi2021wikiasp}, and keywords \cite{wang2021mention, he2020ctrlsum}.
Recently, \citet{shen2022mred} propose a sentence-level labeled meta-review dataset, MReD, for the controllable summarization task from a new control perspective that focuses on controlling the structure of the output summary.
The input consists of several reviews on the same research paper, and a control sequence specifying the desired summary structure. 
For instance, with a control sequence of ``abstract | strength | decision'', the generated output should be composed of a sentence that summarizes the contents of the paper, followed by a sentence discussing the strengths, then the last sentence giving the final decision.


Previous controllable summarization models are commonly fine-tuned on pre-trained transformer architectures \cite{vaswani2017attention} such as BART \cite{lewis2020bart} and Pegasus \cite{zhang2020pegasus}, with the control signals merged into the text input or prompts \cite{shen2022mred, narayan2022well, he2020ctrlsum, keskar2019ctrl, fan2018controllable}.
Previous works mainly focus on improving the summary's similarity with the gold reference, leaving room for further improvement on the controllability.
In particular, the best-performing model on the recently released MReD dataset still generates
around 29\% of the sentences that do not follow the control structure\footnote{
For instance, the generated sentence may discuss the weakness of the paper even though it corresponds to a ``strength'' control label.
}, which is far from satisfactory.

In this paper, we explore how to enhance the structure-controllability in summarization.
Specifically, we notice the following possible pitfalls in the existing summarization models.
First, those models usually treat generation as a standalone process, which continuously generates the tokens solely based on the logits predictions, without stopping to reconsider whether the generated sequences satisfy the control signals.
Moreover, 
autoregressive models can suffer from error propagation in generation due to self-attention
\cite{vaswani2017attention}.
Therefore, if the previous sequences are not well-controlled, subsequent generations may deviate further from the desired output.
Motivated by this, we propose the \textbf{\underline{Sent}}ence-level \textbf{\underline{B}}eam \textbf{\underline{S}}earch (SentBS) method to address the identified issues. For generating each sentence, SentBS first produces multiple sentence options, evaluates and selects the best sentence according to both the control structure as well as the model's log-likelihood, then continues the generation for the next sentence.

Experiments show that SentBS can significantly improve the model's structure-controllability.
In particular, our best setting removes up to 68\% of control mistakes produced by the existing model on MReD without compromising the summarization quality.
The human evaluation further proves that SentBS significantly improves the fluency of summaries.

To summarize, our main contributions are:
(1) To the best of our knowledge, we are the first to conduct sentence-by-sentence controlled generation for structure-controllable summarization.
(2) We propose SentBS, which conducts text generation with continuous evaluation on the sentence level with respect to the control requirements.
  This method can be easily applied to existing autoregressive models.
(3) Experiments show that SentBS significantly increases the model's structure controllability while preserving the summarization quality.

\section{Related Work}
\label{sec:related}

\paragraph{Conditional Text Generation.}
Large pretrained language models have shown impressive performance \cite{lewis2020bart, zhang2020pegasus} on generation tasks.
Many controllable generation tasks \cite{shen2022mred, narayan2022well, chia2022relationprompt, he2020ctrlsum, cheng2020ent, keskar2019ctrl, fan2018controllable} make use of these models by merging the control signals into either the sources or targets.
Our method differs from these approaches by breaking the controlled generation process down to the sentence level, and we make explicit use of the control signals during the generation phase.

\paragraph{Decoding Methods.} 
Common decoding methods include beam search \cite{meister2020best, stahlberg2019nmt, graves2012sequence}, nucleus sampling \cite{holtzman2019curious}, and beam sampling \cite{caccia2019language} (see more details in Appendix \ref{appendic:deocding_strategies}).
In our SentBS method, we use these methods\footnote{\url{https://huggingface.co/docs/transformers/internal/generation_utils}.} as sub-components for generating multiple sentence options.
\citet{hopkins2017automatically}, \citet{ghazvininejad2016generating} and \citet{zhang2014chinese} also works on decoding methods that enforce linguistic constraints on the generation structure.
However, none of these works use a sentence-by-sentence generation and evaluation strategy based on whether the semantic content of the generation meets the control requirements.

\section{Sentence-Level Beam Search}
Given the nature of autoregressive models, errors from previously generated tokens at inference time can be easily propagated to affect the subsequent tokens.
To better control the structure, we propose SentBS, to evaluate and select the outputs at the sentence level during generation. 
SentBS can be easily applied to existing generation models during inference only.
In this section, we explain how SentBS leverages the control sequence where each control label corresponds to the desired sentence category.
Please see more details on the task and dataset in Section \ref{section:task_dataset}.

\subsection{Method Details}

We illustrate the generation process of SentBS in Figure \ref{fig:sentbs}.
Given a generative model, a control sequence consisting of several labels, and the concatenated review texts, SentBS first generates $k$ sentence options in parallel
(e.g., Option 1-0/1-1/1-2/1-3)
using multiple decoding methods such as beam search, nucleus sampling, and beam sampling\footnote{We modify these decoding methods to generate at the sentence level.}.
We calculate a combined score (e.g., Option 1-1 has the best combined score of -0.35) for each sentence
by adding its normalized sequence likelihood with a classifier-predicted probability score of the sentence belonging to the required category (see classifier details in Appendix \ref{appendix:eval_classifier}). 
According to the combined score, we select the top $n$ sentences and feed them individually into the decoder as prompts\footnote{Note that unlike \citet{gehman2020realtoxicityprompts} and \citet{sheng2019woman} that use user-designed prompts, we use the model-generated sentences for continued generation.} for generating the next sentence.



We generate $k$ subsequent sentence options for each of the $n$ prompts, resulting in a total of $k\cdot n$ sentences.
These newly generated sentences will be concatenated after the corresponding prompts to form new prompts for the next generation, and new scores will be calculated to select the next top $n$ prompts.
In particular, the sequence likelihood score is recalculated for the full sequence. 
The same generation process continues until all sentences required in the control sequence are produced.
This generation process is similar to beam search, except that instead of merely selecting tokens based on log-likelihood, we also conduct the selection on the sentence level based on both the control requirements and the sequence likelihood. 

\subsection{Method Applications}
Although SentBS evaluates the structural requirements on a sentence level, it can be easily extended to segment-level control methods 
where each control label corresponds to one or more sentences of the same category. 
In this case, only the first generated sentence has an explicit label, whereas the subsequent sentences can either have the same label as their previous sentences, or carry the next label in the control sequence. 
Therefore, after generating the first sentence, we can again apply SentBS to the generation of subsequent sentences by comparing the probabilities corresponding to the two allowed labels. 
For instance, given a control sequence of “abstract | strength | decision”, after generating the first “abstract” sentence,  we look at the classification score on both “abstract” and “strength” for the second sentence, and assign the label with higher probability to it. 
Thus, each sentence option for the second sentence is assigned a label of either “abstract” or “strength”, depending on the generated content. 
In the same manner, we can generate the subsequent sentences until the model gives the stop signal (i.e., the ``eos'' token). 

Following the logic above, we can further extend SentBS to other controlled generation tasks on a paragraph level (e.g. style transfer), as long as the control requirements can be evaluated on the sentence level. 
For instance, for converting non-formal text to formal text, we can score each sentence according to a combined score of sequence log-likelihood and the degree of formality.
Nevertheless, this is beyond the scope of this paper and we leave the investigation of such applications to future work.

\section{Experiments}

\subsection{Task and Dataset}
\label{section:task_dataset}
We carry out experiments on the benchmark MReD dataset \cite{shen2022mred}. 
Each sentence in the target meta-review is annotated with a category label 
according to its main function from the following:
``abstract'', ``strength'', ``weakness'', ``suggestion'', ``rating summary'', ``rebuttal process'', ``ac disagreement'', ``decision'', and ``misc'' 
(see Appendix \ref{appendix:dataset}).

This dataset comes with two control task settings for structure-controllable summarization: \textit{Sent-Ctrl}, which uses each label in the control sequence to indicate a category for one summary sentence in the summary, and \textit{Seg-Ctrl}, which uses each label in the control sequence to represent a segment of summary sentences of the same category. 
The ultimate goal for both settings is to generate meta-review passages derivable from the reviews while at the same time obeying the required structure.

\begin{table*}[t!]
	\centering
    \scalebox{0.6}{
	    \setlength{\tabcolsep}{1mm}{
        \begin{tabular}{lcccc}
        \toprule
        & Structure$\uparrow$ & Edits $\downarrow$ & BERTScore $\uparrow$ & R$_1$ / R$_2$ / R$_L$ $\uparrow$ \\
        \midrule
        \midrule
        \textbf{Sent-Ctrl} \cite{shen2022mred} &&&& \\
        - Reported & 0.706 & - & - & 38.73 / 10.82 / 23.05 \\
        - Reproduced & 0.737 & 1470.0 & 0.8631 & \textbf{38.97} / \textbf{11.19} / 23.48 \\
        \midrule
        \textbf{Sent-Ctrl + SentBS (ours)} &&&& \\
        Nucleus sampling ($k=4$)  & 0.852 & 785.0 & 0.8627 & 36.50 /  9.11 / 21.76 \\
        Nucleus sampling ($k=5$) & 0.887 & 600.3 & 0.8637 & 36.96 /  9.28 / 22.25 \\
        Nucleus sampling ($k=6$) & 0.891 & 570.3 & 0.8634 & 36.79 /  9.21 / 22.12 \\
        Nucleus sampling ($k=7$) & 0.893 & 567.7 & 0.8642 & 37.00 /  9.34 / 22.30 \\
        Nucleus sampling ($k=8$) & 0.885 & 605.3 & 0.8642 & 37.07 /  9.62 / 22.47 \\ 
        \hdashline[2pt/5pt]
        Beam Sampling ($k=4$) & 0.771 & 1142.0 & 0.8627 & 37.44 / 10.80 / 23.11 \\
        Beam Sampling ($k=5$) & 0.803 & 993.7 & 0.8626 & 37.82 / 10.87 / 23.09 \\
        Beam Sampling ($k=6$) & 0.802 & 998.7 & 0.8621 & 37.71 / 10.86 / 22.93 \\
        Beam Sampling ($k=7$) & 0.795 & 1011.0 & 0.8617 & 37.88 / 10.84 / 22.83 \\
        Beam Sampling ($k=8$) & 0.794 & 1007.0 & 0.8613 & 37.67 / 10.77 / 22.76 \\
        \hdashline[2pt/5pt]
        Beam search + Nucleus sampling ($k=4$) & 0.874 & 665.0 & 0.8644 & 38.23 / 10.47 / 23.22 \\
        Beam search + Nucleus sampling ($k=5$) & 0.887 & 609.7 & 0.8645 & 38.37 / 10.54 / 23.27 \\
        Beam search + Nucleus sampling ($k=6$) & 0.894 & 569.3 & \textbf{0.8648} & 38.36 / 10.67 / 23.35 \\
        Beam search + Nucleus sampling ($k=7$) & 0.904 & 519.3 & \textbf{0.8648} & 38.30 / 10.62 / 23.37 \\
        Beam search + Nucleus sampling ($k=8$) & \textbf{0.915} & \textbf{467.7} & 0.8647 & 38.32 / 10.66 / 23.42 \\
        \hdashline[2pt/5pt]
        Beam search + Beam sampling + Nucleus sampling ($k=4$)  & 0.839 & 833.7 & 0.8645 & 38.33 / 10.96 / 23.48 \\
        Beam search + Beam sampling + Nucleus sampling ($k=5$)  & 0.868 & 700.3 & 0.8644 & 38.39 / 10.95 / \textbf{23.54} \\
        Beam search + Beam sampling + Nucleus sampling ($k=6$)  & 0.883 & 634.0 & 0.8644 & 38.32 / 10.90 / 23.45 \\
        Beam search + Beam sampling + Nucleus sampling ($k=7$)  & 0.885 & 607.7 & 0.8643 & 38.38 / 10.85 / 23.45 \\
        Beam search + Beam sampling + Nucleus sampling ($k=8$)  & 0.893 & 573.0 & 0.8640 & 38.43 / 10.88 / 23.38 \\
        \midrule
        \midrule
        \textbf{Seg-Ctrl} \cite{shen2022mred} &&&& \\
        -Reported & 0.623 & - & - & 36.38 / 10.04 / 21.90 \\
        -Reproduced & 0.755 & 855.0 & \textbf{0.8604} & 36.69 / 10.44 / 22.29 \\
        \midrule
        \textbf{Seg-Ctrl + SentBS (ours)}  & \textbf{0.887} & \textbf{394.3} & 0.8601 & \textbf{36.75 / 10.35 / 22.51} \\
        \midrule
        \bottomrule
        \end{tabular}}}
    \vspace{-2mm}
    \caption{Main results. 
    We divide the table into 2 sections using double horizontal lines.
    The top section shows the Sent-Ctrl baseline and Sent-Ctrl with various settings of SentBS,
    and the bottom section shows the Seg-Ctrl baseline and Seg-Ctrl with SentBS. 
    For the latter section, we use the ``Beam search + Beam sampling + Nucleus sampling'' setting and $k=8$ for SentBS.
    }
    \vspace{-5mm}
	\label{tab:main_results}
\end{table*}

\subsection{Evaluation Metrics}
We evaluate the summarization quality by reporting the F1 scores of the standard Rouge-1/2/L \cite{lin2004rouge} and BERTScore\footnote{\url{https://github.com/Tiiiger/bert_score}} \cite{bert-score}. 
In terms of the structural agreement, we report both the human and automatic evaluation results. 
For the human evaluation, we use ``structure similarity'' following \cite{shen2022mred}, while the automatic evaluation is conducted by passing the summary into a LSTM-CRF classifier \cite{lample2016neural} (see appendix \ref{appendix:structural_eval}).
We also aggregate the total edit distance between the summary structure and the gold structure for the full test set to quantitatively evaluate the control performance. 
This additional metric is named ``total edits''. 

\subsection{Experimental Settings}

We follow \citet{shen2022mred} and reproduce the results using the same reported settings (see Appendix \ref{appendix:set_up}) as baselines.
We use SentBS during inference for \textit{Sent-Ctrl} and \textit{Seg-Ctrl}.
Since \citet{shen2022mred} uses beam search with a beam size of 4 during inference, we set $n=4$ for SentBS to maintain 4 sentences for each generation step as well.
We experiment with different combinations of beam search, nucleus sampling, and beam sampling as decoding sub-components, and explore various $k$ values up to 8 (see more in Appendix \ref{appendix:choice_kn}).
We set \textit{top}-\textit{p} for nucleus sampling (see Appendix \ref{appendic:deocding_strategies}) to 0.9.
All reported results for SentBS are the average of 3 runs 
to mitigate the uncertainty caused by sampling methods.

\subsection{Main Results}
\subsubsection{SentBS for \textit{Sent-Ctrl}}
In Table \ref{tab:main_results}, we first reproduce the ``Sent-Ctrl'' baseline.
We discover that the generated structures hardly follow the corresponding control sequences in a strict sense (see Appendix Table \ref{tab:sent-ctrl}).

Next, we evaluate SentBS with different decoding methods and various $k$ values on the \textit{Sent-Ctrl} model.
Besides SentBS with single decoding methods (\textbf{``Nucleus sampling''} and \textbf{``Beam sampling''}),
we also explore combined decoding methods: 
\textbf{``Beam search + Nucleus sampling''},
where we use beam search to generate 1 sentence option, nucleus sampling for the rest;
and \textbf{``Beam search + Beam sampling + Nucleus sampling''},
where we use beam search
for 1 option, beam sampling for $\frac{k}{2}$ options\footnote{If $\frac{k}{2}$ is not an integer, we use the nearest smaller integer of this value. For instance, if $k=7$, we use beam sampling to generate 3 options.} and nucleus sampling for the rest.

\paragraph{SentBS with single decoding methods.}
In general, both single-method settings
achieves better generation structures as compared to the \textit{Sent-Ctrl} baseline.
For nucleus sampling, the generation quality improves with increasing $k$ values (except for $k=6$), and the structure similarity is reasonably good for $k>4$. 
With more sampled sentences, SentBS has more candidates to choose from to better suit the quality and structural requirements.
However, beam sampling has mediocre structure scores regardless of $k$ values.
This is because beam sampling 
only keeps the top $k$ sampled outputs with the best sequence likelihood.
Therefore, text diversity may suffer and there may not be sufficient sentences of different label categories for SentBS to choose from. 
Nevertheless, neither nucleus sampling nor beam sampling produces sufficiently fluent text as compared to \textit{Sent-Ctrl} (according to Rouge 1/2/L scores).
Nucleus sampling produces much less fluent generations as compared to beam sampling, possibly caused by accidentally sampling some of the less likely tokens which are most probably discarded by beam sampling. 

\paragraph{SentBS with multiple decoding methods.}
Various decoding methods combined can produce a good range of sentence options for both quality and diversity,
so that SentBS can generate a much better structure with good text quality.
In particular, ``Beam search + Nucleus sampling'' consistently achieves the best BertScores under different k values, and has the best structure scores for $k > 5$. 
At $k=8$, more than 68\% of the total edits from \textit{Sent-Ctrl} can be avoided.
``Beam search + Beam sampling + Nucleus sampling'' has better Rouge 1/2/L performances than ``Beam search + Nucleus sampling'', and has the best Rouge-L score when $k=5$.
The essential difference between these two settings is that, for the same $k$ values, the latter replaces some of the former's sentence options from nucleus sampling with beam sampling.
Therefore, it is not surprising that the latter has a slightly worse structure.
Nevertheless, SentBS with either of the combined decoding strategies far exceeds \textit{Sent-Ctrl} in terms of control, while at the same time achieving higher contextual BERTScores and comparable Rouge-L scores, although slightly lower Rouge-1/2 scores.
For an abstractive dataset like MReD, contextual similarity and Rouge-L may be more important metrics to evaluate fluency.

\begin{table}[t!]
	\centering
    \resizebox{1 \columnwidth}{!}{
	    \setlength{\tabcolsep}{1mm}{
        \begin{tabular}{lcccc}
        \toprule
        Beam Size & Structure $\uparrow$ & Edits $\downarrow$ & BERTScore $\uparrow$ & R$_1$ / R$_2$ / R$_L$ $\uparrow$ \\
        \midrule
        4 & \textbf{0.737} & 1470 & \textbf{0.8631} & \textbf{38.97} / \textbf{11.19} / \textbf{23.48} \\
        5 & 0.732 & \textbf{1424} & 0.8625 & 38.81 / 10.92 / 23.34 \\
        6 & 0.723 & 1456 & 0.8618 & 38.67 / 10.85 / 23.19 \\
        7 & 0.727 & 1427 & 0.8613 & 38.55 / 10.73 / 23.00 \\
        8 & 0.722 & 1448 & 0.8608 & 38.29 / 10.68 / 22.90 \\
        \bottomrule
        \end{tabular}}}
    \vspace{-2mm}
    \caption{Beam Search generation results on MReD using Sent-Ctrl with increasing beam sizes. 
    }
	\label{tab:beamsearch_degradation}
	\vspace{-7mm}
\end{table}

\paragraph{Effect of Increasing k Values}
The widely-used beam search method usually experiences worse generation quality with a large beam size.  
On MReD, we also observe that the conventional beam search method experiences consistent decreases in the BertScores and the Rouge 1/2/L scores with increasingly large beam numbers (Table \ref{tab:beamsearch_degradation}).
Nevertheless, it is exciting to see that SentBS exhibits relatively stable or even improving generation quality and structure with increasing $k$ values (Table \ref{tab:main_results}). 
Therefore, it may be easier to apply SentBS to other datasets with less need to tune the $k$ values. 



\begin{table}[t!]
    \footnotesize
	\centering
    \resizebox{0.73\columnwidth}{!}{
	    \setlength{\tabcolsep}{1.5mm}{
        \begin{tabular}{lcc}
        \toprule
        & Sent-Ctrl  & SentBS\\
        \midrule
        Fluency & 0.520 &  0.780*\\
        Content Relevance & 0.680 &  0.700\\
        Structure Similarity & 0.699 &0.838** \\
        Decision Correctness & 0.740 &  0.680\\
        \bottomrule
        \end{tabular}}}
    \vspace{-2mm}
    \caption{Human evaluation. * indicates for p-value $<0.05$, ** for p-value $<0.0001$ by Welsh's t-test.}
    \vspace{-6mm}
	\label{tab:human_eval}
\end{table}

\subsubsection{SentBS for \textit{Seg-Ctrl}}


We also report the results of applying SentBS on the ``Seg-Ctrl'' setting in Table \ref{tab:main_results}. 
For simplification, we show the results of SentBS with the ``Beam search + Beam sampling + Nucleus sampling'' decoding strategy at $k=8$.
Again, SentBS helps to significantly improve the generation structure, while achieving good generation quality.

\subsection{Human Evaluation}
\label{section:human_eval}
We manually evaluate the ``Beam search + Beam sampling + Nucleus sampling'' setting of SentBS at $k=8$ and the \textit{Sent-Ctrl} baseline on 50 random test instances. 
2 Human judges are asked to judge the summary in terms of fluency, content relevance, structure similarity, and decision correctness.
All scores are in the range of 0 to 1 (see Appendix \ref{appendix:human_eval}).

From Table \ref{tab:human_eval}, we can see that SentBS is significantly better than \textit{Sent-Ctrl} in terms of fluency and structure similarity.
SentBS is also more preferred (although not significantly) by human judges for content relevance.
However, it has a lower (not significant) decision correctness.
The model may sample tokens for the wrong decisions, while still satisfying the control requirement.
One way to mitigate this impact is to train the classifier to differentiate various decisions and explicitly use the correct decision in the control sequence.

\section{Conclusions}
In this work, We refocus the attention of controllable summarization from the similarity with gold toward the model's controllability.
We propose a simple and effective method, SentBS, to enhance the structure-controllability of the Transformers BART model trained on MReD. 
With extensive experiments and human evaluation, we show that without the need for retraining, 
SentBS can significantly improve the controllability of summary structure while achieving good generation quality.  

\section*{Limitations}
SentBS requires a combination of maximization-based and sampling-based decoding strategies to work well.
It requires an excessive amount of sentences not used in the final output to be generated.
This wastes a lot of computational resources and makes the generation inefficient.
For instance, the standard beam search for \textit{Sent-Ctrl} requires around 1h to complete.
When we use $k=4$ and $n=4$ for nucleus sampling and beam sampling, the decoding time is extended to around 5h.
For the setting of $k=8$ and $n=4$ for combined decoding methods, the time required for generating the full test set is 16h.
This is also due to the fact that the Huggingface Transformers model does not accept variable-length prompts for parallel decoding, 
so the generation of sentence options for each possible prompt needs to be carried out one by one.
We leave the investigation of the above issues for future work.

So far, we have shown that the performance of SentBS is highly dependent on the nature and combination of the decoding methods.
We will leave the exploration of how to better select and combine decoding methods to future work.

\section*{Acknowledgement}

Yang You is supported by the NUS startup grant, and the Singapore MOE Tier-1 grant.

\bibliography{anthology,custom}
\bibliographystyle{acl_natbib}

\appendix

\clearpage

\section{Decoding Strategies}
\label{appendic:deocding_strategies}
Maximization-based decoding methods such as beam search \cite{meister2020best, stahlberg2019nmt, graves2012sequence} and greedy search (similar to beam search with a beam size of 1) work well for optimizing the generation quality, whereas diversity-based methods such as sampling \cite{holtzman2019curious,fan2018hierarchical,ackley1985learning} focus on improving the generation diversity.
To mitigate the trade-off between generation quality and diversity,
\citet{holtzman2019curious} propose nucleus sampling, where the generator samples only from a nucleus of tokens with cumulative probability larger than a specified $top$-$p$ value, and \citet{caccia2019language} propose beam sampling to integrate sampling with beam search.

\section{Evaluation during Generation}
\label{appendix:eval_classifier}
For evaluating the sequence likelihood, we obtain the log-likelihood of each generated token from the decoder, and then calculate the normalized sequence log-likelihood by averaging the log-likelihood for all tokens in the sequence.
For evaluating the degree to which the generated sentence satisfies the control requirement, we obtain the category log-likelihood for each sentence from a trained classifier.
The trained sentence classifier\footnote{We cannot directly apply the LSTM-CRF tagging model provided with MReD by \citet{shen2022mred}, which requires the full passage in order to output a sequence of labels} is based on the Roberta-Large architecture\footnote{\url{https://huggingface.co/roberta-large}}.
We fine-tune the model using the training set of labeled meta-reviews in the MReD dataset and obtain an accuracy of 0.853 on the test set.

Since both the sequence log-likelihood and classification log-likelihood are derived from probabilities between 0 - 1 and are of the same scale, we didn't experiment with different weightings but simply add them together.
Nevertheless, it would be very straightforward to implement additional weightings to the evaluation process.

\section{The MReD dataset}
\label{appendix:dataset}
Data from the peer review domain has come increasingly popular for research \cite{ hua2019argument, kang2018dataset, cheng2020ape, bhatia2020metagen, bao2021argument, cheng2021argument, bao2022have}.
The MReD dataset consists of 7,089 fully annotated meta-reviews on ICLR 2018-2021 papers with their corresponding reviews.
It is collected from the OpenReview\footnote{\url{https://openreview.net/}} portal.

We use the provided filtered version of MReD dataset\footnote{\url{https://github.com/Shen-Chenhui/MReD/tree/master/summarization/abstractive/filtered_controlled_data}} with the \textit{Sent-Ctrl} and the \textit{Seg-Ctrl} settings for all our experiments.
Specifically, this version contains 5,354 examples for training, 665 examples for validation, and 674 examples for testing. 

The detailed definitions for the categories are as follows:
\squishlist
    \item abstract: A piece of summary about the contents of the submission.
    \item strength: Opinions about the submission’s strengths.
    \item weakness: Opinions about the submission’s weaknesses.
    \item rating summary: A summary about reviewers’ rating scores or decisions
    \item ac disagreement: Area chair (AC) shares different opinions to reviewers.
    \item rebuttal process: Contents related to authors’ rebuttal with respect to reviews or discussions between reviewers in the rebuttal period.
    \item suggestion: Concrete suggestions for improving the submission.
    \item decision: Final decision (i.e., accept or reject) on the submission.
    \item misc: None of the above, such as courtesy expressions.
\squishend

\section{Evaluation of Structural Controllability}
\label{appendix:structural_eval}
For evaluation of structure similarity, \citet{shen2022mred} has manually labeled the sentence category sequence for randomly selected test outputs as the prediction structures.
Treating each category label as a single token, they calculate a normalized edit distance and use 1 to subtract this value for the final similarity score.
Similarly, we use the same method and calculate structure similarity as:
\[ 1 - \frac{edit\_distance(c, r)}{max[len(c), len(r)]} \]
where $r$ stands for the gold reference, $c$ stands for the candidate prediction, and $len(c)$ stands for the total number of labels in the candidate.

In addition, \citet{shen2022mred} has trained a LSTM-CRF\cite{lample2016neural} classifier on the training set's target meta-review passages to predict the sentence categories. 
A high classification accuracy of 0.8583 is reported on the test split, making it possible to evaluate the output structure automatically using this trained classifier.
To do so, we first use the nltk sentence tokenizer to break down the outputs into sentences, then use the above LSTM-CRF sequential tagging model to predict the sentence label sequence.

Nevertheless, the normalized edit distance cannot quantitatively show how frequently the generations disobey the control signals.
Therefore, we count the total edit distance for the whole test set and name this metric ``total edits''.

\section{Experimental Settings}
\label{appendix:set_up}
We reproduce the results of \textit{Sent-Ctrl} and \textit{Seg-Ctrl} by training MReD on the state-of-the-art ``bart-large-cnn'' model\footnote{\url{https://huggingface.co/facebook/bart-large-cnn}}.
Following \citet{shen2022mred}, we set source truncation to 2048, learning rate to 5e-5, and use the Adam optimizer with $\beta_1 = 0.9$, $\beta_2 = 0.999$ and no warm-up nor weight decay.
The decoding method used is beam search with a beam size of 4.
The model is trained for 3 epochs on a single Tesla V100 GPU.

\section{Choice of $k$ and $n$}
\label{appendix:choice_kn}
Inspired by beam search, we explore $k$ up to 8 with $n=4$ for our SentBS since it is similar to beam search to a certain degree.
The Huggingface BART model uses a default beam size of 4 for beam search decoding, and up to 2 times the beam size of tokens are considered during the generation process.

\section{Human Evaluation}
\label{appendix:human_eval}
For fluency and relevance evaluations, the judges are provided with the summaries from the two models in a random order, and asked to give a score of 1 to their preferred generation and 0 to the other. 
They may give both summaries a score of 1 if the two are equally good.

For structure similarity and decision correctness evaluations, we follow \citet{shen2022mred} and ask the judges to predict the control structure and decision from the given summary, which are then evaluated against the gold for the final results.

\begin{table*}[t!]
	\centering
	\scalebox{0.8}{
	    \setlength{\tabcolsep}{1.5mm}{
        \begin{tabular}{p{3cm} p{16cm}}
        \toprule
        \textbf{Control Sequence} & \textbf{Generated Passage} \\
        \midrule
        abstract | strength | strength &
        \textbf{Sent-Ctrl:} \newline
        \textcolor{cyan}{[}This paper proposes a new inductive bias to neural network architecture - namely a divide and conquer strategy know from algorithmics.\textcolor{cyan}{]$\leftarrow$\textsc{abstract}}
        \textcolor{cyan}{[}The reviewers agree that the paper is well written and easy to follow.\textcolor{cyan}{]$\leftarrow$\textsc{strength}}
        \textcolor{red}{[}The experimental section should be extended, as currently the only convincing success story lies in convex hull experiments, and the generalization scores themselves are pretty bad thus it is not clear if this can be called a success story.\textcolor{red}{]$\leftarrow$\textsc{weakness}} \\
        
        & \textbf{SentBS:} \newline
        \textcolor{cyan}{[}This paper proposes a new inductive bias to neural network architecture - namely a divide and conquer strategy know from algorithmics.\textcolor{cyan}{]$\leftarrow$\textsc{abstract}}
        \textcolor{cyan}{[}The paper is well written and clear.\textcolor{cyan}{]$\leftarrow$\textsc{strength}}
        \textcolor{cyan}{[}The proposed approach is novel and interesting.\textcolor{cyan}{]$\leftarrow$\textsc{strength}}
        \\
        \midrule
        abstract | strength | weakness | weakness | weakness | decision &
        \textbf{Sent-Ctrl:} \newline
        \textcolor{cyan}{[}This paper proposes a method for learning branching strategies within branch-and-bound solvers. <sep>\textcolor{cyan}{]$\leftarrow$\textsc{abstract}}
        \textcolor{cyan}{[}Strengths: <sep> - The proposed method is novel and interesting, and the paper is well-written <sep>\textcolor{cyan}{]$\leftarrow$\textsc{strength}}
        \textcolor{cyan}{[}Weaknesses: (1) the experimental setup itself is questionable.\textcolor{cyan}{]$\leftarrow$\textsc{weakness}} 
        \textcolor{red}{[}The authors consider two setups for the SCIP solver, "clean", with depth-first-search node selection and all other functions disabled (whatever that means), and "default".\textcolor{red}{]$\leftarrow$\textsc{weakness}} 
        \textcolor{red}{[}Some experiments are conducted under the "clean" setup (Table 1, Figure 2 and 3) while other experiments are made under "default" setup.\textcolor{red}{]$\leftarrow$\textsc{weakness}}
        \textcolor{cyan}{[}(2) the presented experimental results are inconsistent with the literature, which I find suspicious the overall method description comprises several blind spots and fallacious arguments.\textcolor{cyan}{]$\leftarrow$\textsc{weakness}} 
        \textcolor{cyan}{[}(3) the authors claim their method results in non-myopic policies, without defining what they mean by that. \textcolor{cyan}{]$\leftarrow$\textsc{weakness}} 
        \textcolor{cyan}{[}I agree with the reviewers and recommend rejection of the paper, and encourage the authors to revise the paper based on the reviewers' feedback and resubmit to a different venue. \textcolor{cyan}{]$\leftarrow$\textsc{decision}} \\
        
        & \textbf{SentBS:} \newline
        \textcolor{cyan}{[}The paper proposes a novel framework for learning branching strategies within branch-and-bound solvers based on a simplified graph-convolutional network (GCN) combined with a novelty-search evolutionary strategy (NS-ES) for training, and a new representation of B\&B trees for computing novelty scores.\textcolor{cyan}{]$\leftarrow$\textsc{abstract}}
        \textcolor{cyan}{[}The idea is interesting.\textcolor{cyan}{]$\leftarrow$\textsc{strength}}
        \textcolor{cyan}{[}However, the experimental setup is questionable and the presented experimental results are inconsistent with the literature.\textcolor{cyan}{]$\leftarrow$\textsc{weakness}}
        \textcolor{cyan}{[}In particular, the reviewers found several arguments to be fallacious. \textcolor{cyan}{]$\leftarrow$\textsc{weakness}}
        \textcolor{cyan}{[}The paper is mostly based on the work from Gasse et al., reuses the same code, (almost) the same problem benchmarks, and some portions of text which are identical.\textcolor{cyan}{]$\leftarrow$\textsc{weakness}}
        \textcolor{cyan}{[}Overall, this is a good paper, but not yet ready for publication.\textcolor{cyan}{]$\leftarrow$\textsc{decision}}
        \\
        
        \bottomrule
        \end{tabular}}}
    \caption{Examples of outputs from \textit{Sent-Ctrl}. We use red to highlight locations where generation deviate from the control labels, and cyan otherwise. Note that a <sep> is generated to represent a newline.}
	\label{tab:sent-ctrl}
\end{table*}

\end{document}